\documentclass[letterpaper]{article} %
\usepackage{aaai2026}  %
\usepackage{times}  %
\usepackage{helvet}  %
\usepackage{courier}  %
\usepackage[hyphens]{url}  %
\usepackage{graphicx} %
\usepackage{natbib}  %
\usepackage{caption} %
\usepackage{microtype}
\usepackage{graphicx}
\usepackage{subcaption}
\usepackage{booktabs} % for professional tables
\usepackage{xcolor}
\usepackage{tikz}
\usepackage{tabularx}
\usepackage{ragged2e}
\usepackage[most]{tcolorbox}
\usepackage{placeins}
\usetikzlibrary{backgrounds}

\newcommand{\highlight}[2]{%
  \begin{tikzpicture}[baseline=(X.base)]
    \node[inner sep=0pt] (X) {#2};
    \begin{scope}[on background layer]
      \fill[#1] (X.south west) rectangle (X.north east);
    \end{scope}
  \end{tikzpicture}
}

\usepackage{amsmath}
\usepackage{amssymb}
\usepackage{mathtools}
\usepackage{amsthm}

\usepackage{algorithm}
\usepackage{algorithmic}
\usepackage{booktabs}
\usepackage{newfloat}
\usepackage{listings}
\DeclareCaptionStyle{ruled}{labelfont=normalfont,labelsep=colon,strut=off} %
\floatstyle{ruled}
\newfloat{listing}{tb}{lst}{}
\floatname{listing}{Listing}
\title {D-SCoRE: \textbf{D}ocument-Centric \textbf{S}egmentation and \textbf{Co}T Reasoning \\    with Structured \textbf{E}xport for QA-CoT Data Generation}

\author{
    Weibo Zhou\textsuperscript{\rm1},
    Lingbo Li\textsuperscript{\rm2},
    Shangsong Liang\textsuperscript{\rm3}
}
\affiliations{
    \textsuperscript{\rm 1}City University of Hong Kong,  
    \textsuperscript{\rm2}University of Warwick,
    \textsuperscript{\rm3}Sun Yat-sen University

}

\usepackage{bibentry}

\makeatletter
\renewcommand{\copyright@on}{} %
\makeatother

\begin{document}

\maketitle
\begin{abstract}
The scarcity and high cost of high-quality domain-specific question-answering (QA) datasets limit supervised fine-tuning of large language models (LLMs).
We introduce \textbf{D-SCoRE}, a training-free framework that leverages LLMs and prompt engineering to automatically generate diverse, rich QA datasets with Chain-of-Thought (CoT) from arbitrary textual sources.
By integrating \textbf{D}ocument-centric processing, \textbf{S}egmentation, \textbf{Co}T \textbf{R}easoning, and structured \textbf{E}xport—along with multi-dimensional controls such as semantic role transformation, question type balancing, and counterfactual augmentation—D-SCoRE produces tailored QA pairs with enhanced diversity and relevance.
LLMs fine-tuned on D-SCoRE-generated datasets outperform those trained on human-annotated QA data across most evaluated domains.
Its efficiency and scalability enable rapid, high-performance domain-adaptive fine-tuning on consumer-grade hardware, generating over 1,100 high-quality QA pairs per GPU-hour end-to-end.
\end{abstract}

\section{Introduction}

High-quality question-answering (QA) datasets are essential for supervised fine-tuning (SFT) of large language models (LLMs). However, manually annotated or corpus-extracted datasets (e.g., SQuAD~\cite{rajpurkar-etal-2016-squad}) are expensive, time-consuming, and difficult to scale, often favoring shallow extractive patterns over deep reasoning. This limits rapid domain adaptation and hinders models from learning robust logical inference.

Semi-synthetic approaches improve efficiency but frequently depend on complex preprocessing, pre-annotated seeds, rigid templates, or heavy engineering. They commonly suffer from limited diversity, weak control over question complexity, and shallow logical depth.

To address these shortcomings, we introduce D-SCoRE (Document-centric Segmentation and CoT Reasoning with Structured Export), a training-free, end-to-end pipeline that generates high-fidelity QA-CoT pairs directly from arbitrary texts using carefully designed LLMs prompting. The method combines document-centric segmentation, definite Chain-of-Thought (CoT) reasoning, and counterfactual augmentation to achieve strong factual grounding, semantic diversity, and controllable reasoning depth—without fine-tuning or elaborate infrastructure.

Our main contributions are:
\begin{itemize}
\item A lightweight, fully end-to-end framework that produces QA-CoT pairs from raw text, eliminating complex preprocessing and model training.
\item Novel control mechanisms based on semantic role transformations, step-by-step CoT traces, and counterfactual distractors to improve diversity, complexity, and factual accuracy over previous QA generation methods.
\item Seamless integration of implicit questions with definite reasoning traces, enabling effective reasoning transfer even from extractive-style documents.
\item Empirical results demonstrating that LLMs fine-tuned on D-SCoRE data outperform models trained on human-annotated SQuAD when evaluated on both SQuAD and SQuADShifts benchmarks.
\end{itemize}
Remarkably, D-SCoRE achieves an effective end-to-end throughput of over 1,100 QA pairs per GPU-hour on modern consumer-grade hardware, even when processing typical 100--200-word text segments enriched with CoT reasoning, counterfactual distractors, and question paraphrasing. This enables highly scalable and accessible domain adaptation in resource-constrained environments.

\section{Related Work}
\subsection{QA Dataset Generation}
QA dataset creation approaches fall into two main categories: non-synthetic and semi-synthetic.
Non-synthetic methods (e.g., SQuAD, FairytaleQA~\cite{xu-etal-2022-fantastic}) rely on crowd-sourced questions over existing texts. They offer high relevance and quality but are extremely labor-intensive, expensive, and difficult to scale. Human annotators often produce shallow, extractive questions, limiting models to keyword matching rather than deep reasoning.
Semi-synthetic methods improve efficiency by combining human curation with automation, including answer-first~\cite{alberti-etal-2019-synthetic}, question-first~\cite{RODRIGUEZTORREALBA2022118258}, auxiliary-signal~\cite{10.1145/3366423.3380270}, and LLMs-based techniques such as Self-Instruct~\cite{wang-etal-2023-self-instruct}, data augmentation~\cite{chowdhury-chadha-2024-generative}, optimized pipelines~\cite{yuen2025automaticdatasetgenerationknowledge}, and structured reasoning recipes~\cite{guha2025openthoughtsdatarecipesreasoning}. However, most still depend on complex preprocessing, pre-annotated seeds, rigid templates, or heavy engineering, falling short of true end-to-end automation.
Existing methods rarely provide explicit guidance toward intrinsic reasoning patterns, motivating simpler, training-free frameworks that generate high-fidelity QA with CoT pairs directly from raw text with strong grounding and minimal infrastructure.
\subsection{QA Data Quality Enhancement}
Quality improvement efforts focus on accuracy, richness, and diversity.
Accuracy is enhanced via implicit relation extraction~\cite{DBLP:journals/corr/abs-1905-07471}, summarization~\cite{dugan-etal-2022-feasibility}, or structured planning~\cite{li-zhang-2024-planning}, though often at high computational cost or limited generality.
Richness is increased through domain attributes~\cite{xu-etal-2022-fantastic}, distractors, multiple-choice formats~\cite{RODRIGUEZTORREALBA2022118258}, or additional annotations, yet semantic depth is rarely addressed.
Diversity—critical for robust models~\cite{sultan-etal-2020-importance}—is pursued via stochastic decoding~\cite{yadav-etal-2024-explicit}, entity substitution~\cite{paranjape-etal-2022-retrieval}, position/category conditioning~\cite{yadav-etal-2024-explicit}, or iterative refinement~\cite{eo-etal-2023-towards}. These approaches frequently suffer from poor semantic control, redundancy, fluency issues, or architectural complexity.
Most methods still generate isolated QA pairs without explicit reasoning traces, limiting their value for multi-step inference training. D-SCoRE addresses this by natively producing questions together with step-by-step CoT reasoning, shifting focus from quantity scaling to genuine reasoning quality improvement.
\subsection{Downstream Applications}
QA datasets support LLMs fine-tuning (e.g., LoRA on reasoning benchmarks~\cite{guha2025openthoughtsdatarecipesreasoning}), evaluation (MMLU~\cite{DBLP:journals/corr/abs-2009-03300}, GSM8K~\cite{DBLP:journals/corr/abs-2110-14168}, HotpotQA~\cite{yang-etal-2018-hotpotqa}, etc.), and real-world use cases (education~\cite{xu-etal-2022-fantastic}, healthcare~\cite{bai2024hardquestionssyntheticdata}, conversational AI).
However, high curation costs, slow updates, limited diversity, domain specificity, and persistent biases hinder scalable, general-purpose adaptation to new tasks and domains.
These gaps highlight the need for lightweight, versatile, high-quality QA-CoT generation methods that enable efficient domain adaptation with low resource demands.

\section{Methodology}

\begin{figure}[h!]
\centering
\includegraphics[width=0.95\columnwidth]{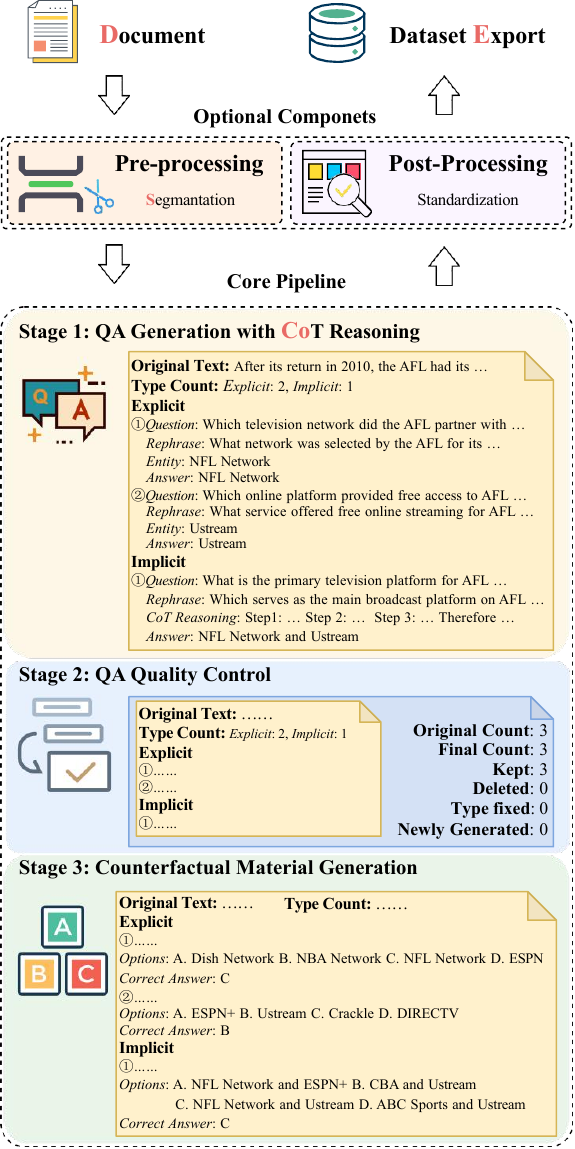}
\caption{Overview of D-SCoRE framework: an LLM-driven three-stage process (QA generation, quality control, counterfactual augmentation) with multi-dimensional control over question types and reasoning depth, plus optional pre-/post-processing components.}
\label{fig:d-score_overview}
\end{figure}

\subsection{Core framework}

The core framework comprises three stages where an LLM generates or refines QA pairs. Figure~\ref{fig:d-score_overview} overviews the framework, including the stages, multi-dimensional control mechanisms, and optional pre-processing components. The prompt templates and the implementation details of the quality control and regeneration logic is provided in appendix and supplementary material.

\subsubsection{Stage 1: QA Generation}
In this stage, QA pairs are synthesized from text segments with an emphasis on heterogeneity through a balanced mix of explicit and implicit questions: explicit questions target directly retrievable factual elements, while implicit questions require inferential reasoning or synthesis of latent insights across sentences. Via targeted prompt engineering, the LLM is instructed to identify discrete entities or informational units for explicit questions, perform cross-sentence inferential reasoning and integrate latent insights for implicit questions, and enhance question variability through semantic role permutations while preserving answer invariance. Unlike traditional explicit-focused annotations, D-SCoRE prioritizes implicit questions augmented with CoT traces, enabling models to develop reasoning capabilities beyond surface-level matching.

\textbf{Formalization:} To guide generation and enable effective reasoning transfer, we formally distinguish between explicit and implicit questions. A question $Q$ is \textit{explicit} if its answer $A$ can be directly extracted as a verbatim span from the source text $T$:
\begin{align*}
Q_{\text{explicit}} &\iff \exists \, s \subseteq T \text{ s.t. } A = \text{extract}(s) \\
&\quad \text{and } Q \text{ queries } s \text{ verbatim},
\end{align*}
where $s = T[i:j]$ is a contiguous subsequence ($0 \leq i < j \leq |T|$).

Conversely, a question $Q$ is \textit{implicit} if $A$ requires reasoning across multiple spans or relationships in $T$:
\begin{align*}
Q_{\text{implicit}} &\iff \forall \, s \subseteq T, \, A \neq \text{extract}(s) \\
&\quad \text{or } Q \text{ infers from multiple } s_k.
\end{align*}

This distinction is central to D-SCoRE. Explicit questions encourage surface-level matching, while implicit questions—augmented with CoT traces—compel the model to internalize deeper semantic relationships and reasoning patterns~\cite{NEURIPS2022_9d560961}.

\subsubsection{Stage 2: QA Quality Control}
To ensure the precision and relevance of the synthesized QA pairs, this stage implements rigorous quality assurance through two key verifications. First, textual fidelity is checked by verifying that each QA pair is exclusively grounded in the source text, with no hallucinations or incorporation of extraneous knowledge. Second, interrogative taxonomy validation classifies each question as explicit or implicit, confirming that explicit questions address directly extractable information while implicit questions genuinely require reasoning or synthesis.

Misclassified or inconsistent pairs (e.g., implicit questions with verbatim answers or lacking inference basis) are rectified or regenerated. Each segment must yield at least one explicit question. Prompt engineering enables iterative refinement of aberrant pairs, preserving implicit question purity for reasoning transfer and enhancing dataset integrity.

Formalization: We define the quality of a QA pair as the product of two key indicators:
\begin{align*}
\text{Quality}(Q) &= \text{Fidelity}(Q, A, T) \times \text{Taxonomy}(Q),
\end{align*}
where $  \text{Fidelity}(Q, A, T) = 1  $ if the answer $  A  $ is fully and faithfully grounded in the source text $  T  $ without hallucinations (0 otherwise), and $  \text{Taxonomy}(Q) = 1  $ if the question $  Q  $ correctly adheres to the explicit/implicit taxonomy defined in Stage 1 (0 otherwise). While this presents a binary formulation for conceptual clarity, in practice we employ a multi-step verification process to approximate this quality function. Only pairs estimated to achieve $  \text{Quality}(Q)=1  $ are retained. The Critic model (a heterogeneous Stage 2 model, e.g., DeepSeek-R1) serves as the core mechanism for minimizing $  D_{\text{KL}}  $ divergence between synthetic and ideal distributions, enforcing high-quality filtering and preventing the accumulation of synthetic noise. Full prompt templates and quality control implementation details are provided in appendix and supplementary material.

\subsubsection{Stage 3: Counterfactual Material Generation}
To increase dataset sophistication and enhance SFT effectiveness across diverse domains, this final stage generates counterfactual alternatives for each QA pair. Guided by prompt engineering, the LLM synthesizes three distractor alternatives per QA pair—ensuring semantic proximity to the correct answer while introducing factually erroneous yet plausible content grounded in the source text—evaluates their quality in terms of contextual alignment and discriminability (regenerating subpar distractors as needed), and randomizes the position of the correct answer among the four options to mitigate positional biases. At least one distractor must exploit nuanced semantic deviations to enhance complexity for robust SFT.

Formalization: For each QA pair $  (Q, A)  $, we generate three counterfactual distractors $  D = \{d_1, d_2, d_3\}  $ satisfying the following heuristic constraints via prompt engineering: semantic proximity (distractors close to $  A  $ but incorrect), factuality grounding (derived as modified spans from $  T  $, ensuring credibility without hallucinations), and discriminability (sufficiently distinct from each other to encourage robust decision-making). These constraints are enforced through carefully designed prompts that instruct the LLM to produce plausible yet erroneous alternatives. This approach draws inspiration from plausibility-focused datasets such as PlausibleQA~\cite{Mozafari_2025} and supports improved model robustness by training on challenging near-miss options.

\subsection{Reasoning-Centric Supervision}

D-SCoRE prioritizes implicit questions paired with CoT traces. This design draws on prior work showing that training with explicit step-by-step reasoning enhances semantic inference and reduces dependence on surface-level matching~\cite{NEURIPS2022_9d560961}.

By emphasizing implicit QA-CoT pairs, extractive texts become reasoning-rich supervision signals, supporting better generalization—even on extractive downstream tasks—through reasoning transfer.

Table~\ref{tab:qualitative-info-content} qualitatively compares the information content across QA types. Here, $I(R; Q, A)$ denotes the mutual information between the reasoning trace $R$ and the question-answer pair $(Q, A)$, reflecting how much additional semantic signal the reasoning process provides. Similarly, $H(A \mid Q, D)$ represents the conditional entropy of the correct answer given the question and distractors, indicating decision difficulty in multiple-choice settings. The table shows that implicit questions with CoT offer substantially richer training signals than explicit ones, while counterfactual formats add controlled robustness at moderate cost. These qualitative distinctions motivate our prioritization of high implicit ratios and CoT augmentation.

\begin{table}[t]
\centering
\scriptsize
\caption{Qualitative Comparison of Information Content Across QA Types: Mutual Information $I(R; Q, A)$ and Conditional Entropy $H(A \mid Q, D)$.}
\label{tab:qualitative-info-content}
\setlength{\tabcolsep}{1.5pt} 
\setlength{\extrarowheight}{1.2pt} 
\begin{tabular}{%
    >{\raggedright\arraybackslash}m{30pt}%
    >{\raggedright\arraybackslash}m{90pt}%
    >{\raggedright\arraybackslash}m{50pt}%
    >{\raggedright\arraybackslash}m{50pt}%
}
\toprule
\textbf{QA Type} & \textbf{Information Content} & \textbf{Benefit to Fine-Tuning} & \textbf{Potential Limitation} \\
\midrule
Explicit & Low ( $ I(R;Q,A) \approx 0 $ ) & Supports basic extraction tasks & Limited transfer to complex reasoning \\
Implicit with CoT & High ( $ I(R;Q,A) \gg 0 $ ) & Enables strong reasoning transfer & Risk of incoherent traces in small models \\
With Counterfactuals & Medium ( $ H(A \mid Q,D) \geq \log 4 - \epsilon $ ) & Improves robustness & Increased generation overhead \\
\bottomrule
\end{tabular}
\end{table}

\subsection{Optional Components}

\subsubsection{Pre-Processing}

In light of LLMs' constrained contextual apertures, an elective pre-processing phase partitions input texts into manageable segments to expedite QA synthesis. This adjunctive procedure accommodates variegated texts via adjustable length parameters aligned with bespoke requisites, thereby modulating question density in synergy with the core framework. For structured corpora, segmentation exploits innate delineations such as chapters or sections. For amorphous texts, entrenched semantic-preserving chunking methodologies, exemplified by Meta-Chunking~\cite{zhao2024metachunking}, are enlisted to retain conceptual coherence. Albeit peripheral to this inquiry, these techniques underpin dataset malleability.

\subsubsection{Post-Processing}

An elective post-processing phase polishes the synthesized dataset by scrutinizing QA pair coherence, excising redundancies, sieving formatting aberrations, or normalizing for SFT via schema validation or LLMs invocations. This phase adapts to idiosyncratic applications, ensuring a refined deliverable for ensuing endeavors.

\section{Experiments}

In this section, we delineate the experimental framework for assessing the D-SCoRE framework across heterogeneous domains, incorporating the SQuAD and SQuADShifts datasets alongside detailed implementation particulars. The experiments are orchestrated to systematically interrogate a sequence of research questions that progressively scrutinize the baseline efficacy, reasoning transfer effects, and operational efficiency of our paradigm in synthesizing QA-CoT datasets for SFT of LLMs.

\subsection{Research Questions}
We address the following research questions to assess D-SCoRE.

\begin{itemize}
\item \textbf{RQ1 (Efficacy vs. Gold Baseline):} Compared to human-annotated gold-standard QA data, how does D-SCoRE's synthesized data perform on in-distribution and out-of-distribution tasks?
\item \textbf{RQ2 (Impact of Implicit-Explicit Ratio):} How does the ratio of implicit (reasoning-intensive) to explicit (extractive) data affect model performance? Does a higher proportion of implicit data enable reasoning transfer, improving performance on extractive benchmarks?
\item \textbf{RQ3 (Model Scaling Effect):} Does D-SCoRE provide consistent benefits across models of varying parameter scales?
\item \textbf{RQ4 (Heterogeneous Quality Control):} Does employing a distinct model for Stage 2 (quality filtering and counterfactual generation) improve the fidelity of generated QA-CoT data and downstream SFT performance compared to a fully homogeneous pipeline?
\end{itemize}

\subsection{Experimental Setup}

\subsubsection{Datasets and Evaluation Metrics}
We harness a SQuAD derivative from QG-Bench~\cite{ushio-etal-2022-generative}, comprising QA pairs derived from 16,462 textual excerpts (each 100--200 words) in the SQuAD training set. These serve as gold-standard references for SFT and as evaluative baselines, while the underlying raw texts function as input for D-SCoRE generation. For evaluation, we adopt the QG-Bench-adapted SQuAD validation set and its SQuADShifts extensions~\cite{pmlr-v119-miller20a}, encompassing four domain-specific subsets (Amazon, NYT, New Wiki, Reddit) to assess in-domain and cross-domain generalization. We evaluate using a comprehensive suite of metrics including F1, Exact Match (EM), and semantic similarity scores (e.g., BLEURT, ROUGE); due to space constraints, only F1 and EM are reported in the main text.

\subsubsection{Data Compositions}
To investigate the impact of implicit-to-explicit ratios (RQ2) under controlled training scale, we conducted experiments on fixed-size subsets randomly sampled from the full generated pool. The standard training budget was set to 22,446 QA pairs per configuration, derived as approximately 5/6 of the maximum available implicit data pool for consistency across ratios; all subsets were randomly sampled without manual curation. The 120\% implicit variant uses the full implicit data pool at 26,936 pairs (1.2× the standard budget), allowing us to probe the performance upper bound under maximally reasoning-centric supervision.

The resulting compositions are summarized in Table~\ref{tab:data-compositions}.

\begin{table}[t]
\centering
\scriptsize
\setlength{\tabcolsep}{5pt} 
\caption{Data composition for fine-tuning with varying ratios of implicit-reasoning QA pairs. All configurations (except 120\% implicit) use exactly 22,446 QA pairs randomly sampled from the generated pool.}
\label{tab:data-compositions}
\begin{tabular}{cccccccc}
\toprule
Ratio & 0\% & 20\% & 40\% & 60\% & 80\% & 100\% & 120\% \\
\midrule
Imp.   & 0       & 3,741   & 7,482   & 11,223  & 14,964  & 18,705  & 26,936 \\
Exp.   & 22,446  & 18,705  & 14,964  & 11,223  & 7,482   & 3,741   & 0      \\
Total  & 22,446  & 22,446  & 22,446  & 22,446  & 22,446  & 22,446  & 26,936 \\
\bottomrule
\end{tabular}
\end{table}

\begin{figure}[t]
\centering
\includegraphics[width=0.95\columnwidth]{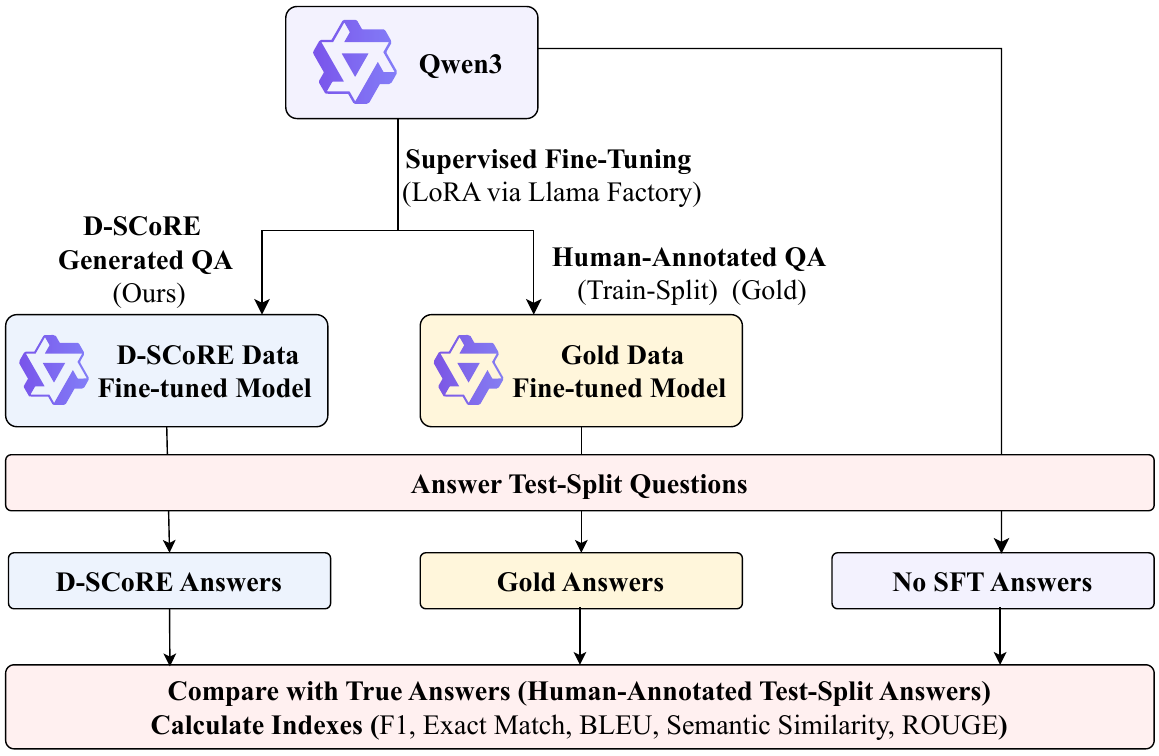}
\caption{Experimental setup for evaluating D-SCoRE-generated QA data via SFT and downstream performance comparison.}
\label{fig:experiment_overview}
\end{figure}

\subsubsection{Implementation Details}

\paragraph{Data Generation}
The D-SCoRE framework was configured to generate 2 explicit and 1 implicit QA-CoT pair per textual segment in Stage 1. Stage 2 performed fidelity checking and explicit/implicit taxonomy validation (with KEEP/DELETE/TYPEFIX and regeneration on failure). Stage 3 produced 3 counterfactual distractors per QA pair with randomized correct-answer positions, followed by question paraphrasing for diversity. This setup produced a full pool of 69k explicit and 27k implicit QA-CoT pairs (97k before paraphrasing), enabling fair-scale comparison to the human-annotated SQuAD Gold dataset (75k pairs).

Two generation modes were compared to evaluate framework homogeneity (RQ4):
\begin{itemize}
\item \textbf{Heterogeneous}: Stage 1 used Qwen3-8B; Stages 2--3 used DeepSeek-R1-7B.
\item \textbf{Homogeneous}: All stages used Qwen3-8B.
\end{itemize}

Generation was conducted on a single NVIDIA RTX 5090 GPU (32GB VRAM) under Linux, with models loaded via Ollama or LM Studio.

\paragraph{Fine-Tuning}
We fine-tuned Qwen3-4B and Qwen3-8B using LoRA (rank=8, alpha=16, learning rate 1e-4) on each data composition. SFT was performed on a single NVIDIA A100 GPU (80GB).

\paragraph{Computational Cost and Efficiency}
We measured the actual generation cost of the D-SCoRE framework on an RTX 5090, processing 2,058 segments through the full framework took 632 minutes (10.53 GPU-hours), yielding 12,262 final QA pairs after paraphrasing. This corresponds to an effective throughput of $ \approx $19.4 QA pairs per minute, or $ \approx $1,164 pairs per GPU-hour. These results highlight D-SCoRE's high efficiency and scalability on consumer-grade hardware for generating large-scale, reasoning-rich QA datasets.

All prompts and framework implementation details are released in the supplementary material. The complete experimental workflow is shown in Figure~\ref{fig:experiment_overview}.

\subsection{Results Analysis}

We evaluate the performance of Qwen3-4B and Qwen3-8B models fine-tuned on various D-SCoRE data compositions using F1 and EM metrics. The empirical outcomes are summarized in the following sections, systematically addressing the research questions.

Figure~\ref{fig:experiment result} visualizes F1 and EM trends across implicit-explicit ratios for both model scales on each evaluation set. Blue lines represent EM scores, orange lines represent F1 scores, with dashed horizontal lines indicating gold and nosft baselines. The plots reveal consistent patterns: performance generally improves with higher implicit ratios, often surpassing gold baselines, demonstrating D-SCoRE's efficacy in synthesizing reasoning-rich data.

Due to space constraints, the main text reports only F1 and EM scores (Figure~\ref{fig:experiment result}) and selected semantic metrics (Table~\ref{tab:semantic_metrics}). Full results across all evaluation metrics and model configurations are provided in the appendix.

\subsubsection{RQ1: Efficacy vs. Gold Baseline}
D-SCoRE-generated training data achieves performance that is competitive with, and in many cases surpasses, the human-annotated SQuAD Gold baseline on both in-distribution and out-of-distribution benchmarks, while using only approximately one-third of the training examples (22,446–26,936 vs. 75,722). This demonstrates a substantial improvement in data efficiency relative to large-scale human-annotated extractive data.

\begin{figure}[h!]
\centering
\includegraphics[width=\columnwidth]{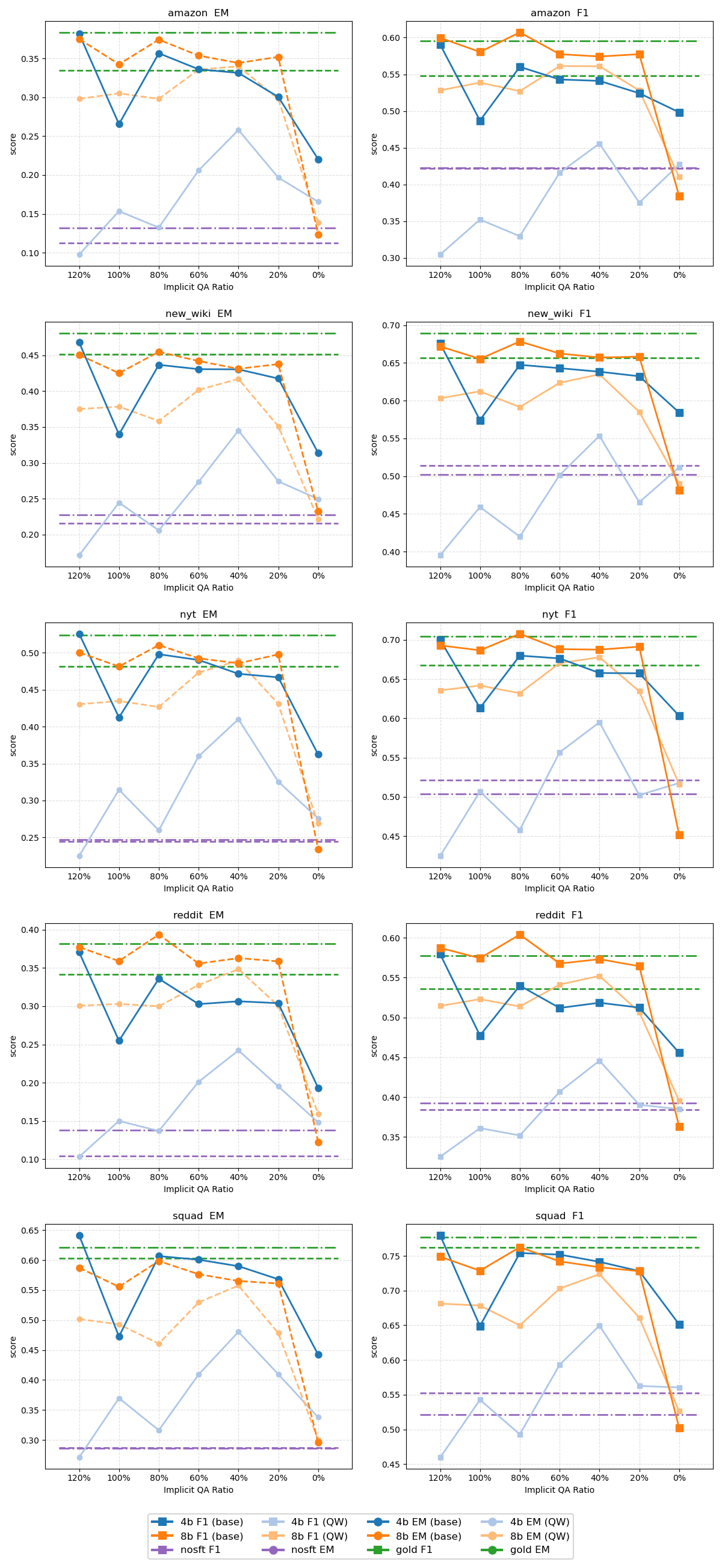}
\caption{F1 and EM performance of Qwen3-4B and Qwen3-8B models fine-tuned on D-SCoRE data with varying implicit--explicit ratios (0\%, 20\%, ..., 120\% implicit) across datasets. Blue/orange lines denote EM/F1 scores; dashed horizontal lines indicate the \textit{gold} (human-annotated) and \textit{nosft} (no fine-tuning) baselines.}
\label{fig:experiment result}
\end{figure}

As shown in Figure~\ref{fig:experiment result}, models fine-tuned on D-SCoRE data consistently narrow the performance gap to Gold across all evaluation domains, and outperform the Gold baseline on multiple datasets and metrics—particularly on out-of-distribution benchmarks under high-implicit configurations. These trends hold for both Qwen3-4B and Qwen3-8B models, indicating robustness across model scales.

\textbf{Key insight:} D-SCoRE data matches or surpasses the SQuAD Gold on both in-distribution and out-of-distribution tasks using only 1/3 the training examples, showing superior data efficiency and cross-domain generalization.

\subsubsection{RQ2: Impact of Implicit-Explicit Ratio and Reasoning Transfer}
Model performance exhibits a clear and consistent improvement as the proportion of implicit (reasoning-intensive) QA-CoT data increases. As shown in Figure~\ref{fig:experiment result}, both F1 and EM scores rise monotonically as the implicit ratio increases from 0\% to 120\% across all five evaluation domains and both model scales.

Training with purely extractive synthetic data (0\% implicit) yields the weakest performance among D-SCoRE configurations and often underperforms the Gold baseline. In contrast, configurations dominated by implicit QA-CoT pairs (80\%–120\%) achieve the strongest results, even though all evaluation benchmarks remain extractive in nature. This demonstrates that reasoning-intensive supervision transfers effectively to extractive downstream tasks.

Interestingly, the 80\% implicit configuration often matches or slightly exceeds the 120\% implicit setting, suggesting that a predominantly implicit yet partially grounded implicit–explicit mixture can achieve near-optimal performance. This indicates diminishing returns when implicit supervision completely replaces extractive signals.

These trends provide strong empirical support for the reasoning transfer decomposition introduced in Section 3.2. As the implicit ratio  \( r \) increases, the contribution of reasoning supervision  \( \Delta_{\text{reasoning}} \) becomes dominant, enabling models to leverage the additional mutual information  $I(R; Q, A)$ encoded in COT traces. The observed monotonic gains confirm that this reasoning signal is effectively internalized and generalized beyond the training format.

\textbf{Key insight:} Higher proportions of implicit QA-CoT data drive strong reasoning transfer to extractive benchmarks, yielding monotonic performance gains and superior results compared to purely extractive supervision.

\subsubsection{RQ3: Model Scaling Effect}
The benefits of high-implicit D-SCoRE data are consistent and robust across model scales.

The 8B model achieves higher absolute performance than the 4B model, with larger gains on challenging out-of-distribution domains (e.g., Reddit at 80\% implicit: 60.41 vs. 53.99 F1). Both scales show the same trajectory: monotonic improvement with increasing implicit ratios, peaking at 80\%–120\%.
Relative gains over Gold are comparable or greater for the 4B model, highlighting D-SCoRE's value in resource-constrained settings where reasoning supervision offsets capacity limitations.

\textbf{Key insight:} D-SCoRE's high-implicit data composition delivers consistent benefits across model scales (4B to 8B), with predictable monotonic gains and particularly pronounced relative improvements for smaller models.

\subsubsection{RQ4: Heterogeneous Collaboration in Quality Control}
The heterogeneous pipeline (Qwen3-8B for Stage 1; DeepSeek-R1-7B for Stages 2–3) significantly outperforms the homogeneous pipeline (Qwen3-8B throughout), as shown by large positive $ \Delta $F1 and $ \Delta $EM values (heterogeneous minus homogeneous) in Table~\ref{tab:implicit_ratio}.

For the 4B model, heterogeneous gains are dramatic: $ \Delta $F1 up to 32.02\% and $ \Delta $EM up to 37.06\% on SQuAD at 120\% implicit ratio, with consistent 18\%–28\% $ \Delta $F1 across all domains in high-implicit regimes (80\%–120\%).

For the 8B model, gains are more moderate but persistent ($ \Delta $F1 5\%–11\%, ~6\%–7\% at 120\% implicit), indicating that even stronger models cannot fully compensate for the quality deficits of homogeneous data.

These gains validate the theoretical justification for iterative refinement in Section 3.1.2, where heterogeneous collaboration enhances Quality(Q) by reducing synthetic noise and hallucinations through superior taxonomy validation and fidelity checks.

Experimental data show that the 4B model is highly sensitive to heterogeneous collaboration, indicating that smaller-scale models are particularly vulnerable to logical flaws in synthetic data without external high-quality reasoning model supervision. The homogeneous pipeline’s severe performance collapse at high implicit ratios ($\Delta$F1 up to 32.02\%) demonstrates that, without a high-performance reasoning model for supervision, synthetic reasoning data tends to amplify errors rather than improve performance. Replacing Stage 2 with a distinct model is therefore a scientific necessity for maintaining data fidelity—minimizing $  D_{\text{KL}}  $ divergence—beyond mere engineering optimization.

\textbf{Key insight:} Employing a distinct high-quality Critic model for Stage 2 substantially improves data fidelity and downstream SFT performance compared to a fully homogeneous pipeline, preventing error amplification in high-implicit regimes and enabling synthetic QA-CoT data to surpass human-annotated standards.

\begin{table}[t]
\centering
\scriptsize
\setlength{\tabcolsep}{1pt}
\caption{Performance improvement of heterogeneous over homogeneous pipeline ($  \Delta  $ F1 /  $  \Delta  $ EM) across implicit ratios and model scales. Values represent $ \Delta $  = heterogeneous -- homogeneous. Yellow/gray highlight indicates highest/second-highest improvement within each model scale.}
\label{tab:implicit_ratio}
\begin{tabular}{@{}cccccc@{}} 
\toprule
\textbf{Imp. \%} & \textbf{Amazon} & \textbf{New Wiki} & \textbf{NYT} & \textbf{Reddit} & \textbf{SQuAD} \\
\midrule
\multicolumn{6}{c}{\textbf{4B} ($  \Delta  $ F1 /  $  \Delta  $ EM)} \\
\midrule
0\%    & 7.11 / 5.49   & 7.26 / 6.45   & 8.56 / 8.65   & 7.08 / 4.51   & 9.09 / 10.47 \\
20\%   & 14.88 / 10.38 & 16.69 / 14.36 & 15.5 / 14.11  & 12.21 / 10.89 & 16.55 / 15.85 \\
40\%   & 8.52 / 7.34   & 8.51 / 8.54   & 6.28 / 6.14   & 7.29 / 6.43   & 9.18 / 10.96 \\
60\%   & 12.70 / 13.01  & 14.16 / 15.74 & 11.95 / 13.02 & 10.52 / 10.16 & 15.87 / 19.15 \\
80\%   & \highlight{gray!30}{23.06}/ \highlight{gray!30}{22.44} & \highlight{gray!30}{22.77}/ \highlight{gray!30}{23.13} & \highlight{gray!30}{22.19}/ \highlight{gray!30}{23.81} & \highlight{gray!30}{18.81}/ \highlight{gray!30}{19.94} & \highlight{gray!30}{26.12}/ \highlight{gray!30}{29.05} \\
100\%  & 13.47 / 11.21 & 11.51 / 9.52  & 10.64 / 9.78  & 11.63 / 10.48 & 10.64 / 10.27 \\
120\%  & \highlight{yellow}{28.59}/ \highlight{yellow}{28.43} & \highlight{yellow}{28.07}/ \highlight{yellow}{29.73} & \highlight{yellow}{27.54}/ \highlight{yellow}{30.04} & \highlight{yellow}{25.44}/ \highlight{yellow}{26.77} & \highlight{yellow}{32.02}/ \highlight{yellow}{37.06} \\
\midrule
\multicolumn{6}{c}{\textbf{8B} ($  \Delta  $ F1 /  $  \Delta  $ EM)} \\
\midrule
0\%    & -2.59 / -1.54 & -0.91 / 1.06  & -6.39 / -3.44 & -3.22 / -3.70  & -2.39 / -0.43 \\
20\%   & 4.93 / 5.40    & \highlight{gray!30}{7.32}/ \highlight{gray!30}{8.69}  & 5.68 / 6.70    & 5.76 / 5.81   & 6.68 / 8.27 \\
40\%   & 1.32 / 0.43   & 2.24 / 1.41   & 0.97 / -0.38  & 2.10 / 1.45    & 1.00 / 0.77 \\
60\%   & 1.62 / 1.80    & 3.88 / 4.03   & 1.77 / 1.92   & 2.65 / 2.79   & 3.91 / 4.69 \\
80\%   & \highlight{yellow}{8.00}/ \highlight{gray!30}{7.62}      & \highlight{yellow}{8.70}/ \highlight{yellow}{9.65}   & \highlight{yellow}{7.57}/ \highlight{yellow}{8.36}   & \highlight{yellow}{9.03}/ \highlight{yellow}{9.39}  & \highlight{yellow}{11.27}/ \highlight{yellow}{13.79} \\
100\%  & 4.17 / 3.75   & 4.31 / 4.69   & 4.48 / 4.67   & 5.14 / 5.61   & 5.02 / 6.27 \\
120\%  & \highlight{gray!30}{7.16}/ \highlight{yellow}{7.71}  & 6.87 / 7.55   & \highlight{gray!30}{5.73}/ \highlight{gray!30}{7.00}     & \highlight{gray!30}{7.29}/ \highlight{gray!30}{7.65} & \highlight{gray!30}{6.76}/ \highlight{gray!30}{8.56} \\
\bottomrule
\end{tabular}
\end{table}

\subsubsection{Semantic Quality and Domain Adaptability}

We evaluate generative performance using BLEU, Semantic Similarity (SemSim), ROUGE-1, ROUGE-2, and ROUGE-L to complement F1 and EM. For brevity, we report results only on NYT (news) and Reddit (conversational discourse)—domains that demand high semantic fluency—in Table~\ref{tab:semantic_metrics}; full results are in the appendix.

The 120\% and 80\% implicit configurations consistently outperform other settings, achieving higher SemSim and ROUGE-L scores and frequently surpassing the Gold standard in ROUGE-2, particularly on NYT. The observed high ROUGE-L and SemSim scores validate the diversity enhancement via semantic role permutations (Section 3.3.2), contributing to strong domain adaptability.

This robust performance across stylistically distinct yet fluency-sensitive domains highlights the effectiveness of D-SCoRE's diversity-focused strategies in producing semantically rich QA-CoT data.

\begin{table}[t]
\centering
\scriptsize
\setlength{\tabcolsep}{1.2pt}
\caption{Performance of 4B and 8B models across datasets and implicit ratios on generation metrics. Each cell shows 4B / 8B scores; within each dataset and metric, top two values for each model size are highlighted in yellow (highest) and gray (second-highest).}
\begin{tabular}{@{}cccccc@{}}
\toprule
Config & BLEU $\uparrow$ & SemSim $\uparrow$ & ROUGE-1 $\uparrow$ & ROUGE-2 $\uparrow$ & ROUGE-L $\uparrow$ \\
\midrule
\multicolumn{6}{c}{\textbf{NYT ( 4B / 8B)}} \\
\midrule
NOSFT & 11.48 / 12.89 & 65.66 / 64.23 & 51.80 / 50.82 & 33.79 / 32.71 & 51.52 / 50.50 \\
\cmidrule(lr){1-6}
0\% & 17.52 / 12.40 & 71.89 / 62.37 & 60.81 / 45.71 & 39.98 / 30.12 & 60.57 / 45.45 \\
20\% & 19.36 / 22.57 & 77.07 / 78.45 & 67.02 / 70.22 & 41.83 / \highlight{gray!30}{45.35} & 66.72 / 69.96 \\
40\% & 19.83 / 22.18 & 77.45 / 78.36 & 67.18 / 69.87 & 42.03 / 44.93 & 66.90 / 69.61 \\
60\% & 19.94 / 22.35 & 78.67 / 78.52 & 69.11 / 69.99 & \highlight{gray!30}{42.72}/ 45.02 & 68.80 / 69.70 \\
80\% & 20.16 / \highlight{yellow}{23.40} & 78.95 / \highlight{gray!30}{79.95} & 69.67 / \highlight{gray!30}{72.01} & 43.36 / \highlight{yellow}{46.43} & 69.39 / \highlight{gray!30}{71.74} \\
100\% & 17.64 / 22.60 & 73.50 / 78.50 & 62.64 / 70.01 & 39.34 / 45.06 & 62.31 / 69.71 \\
120\% & \highlight{yellow}{21.26}/ 23.32 & \highlight{gray!30}{80.44}/ 78.82 & \highlight{yellow}{71.64}/ 70.48 & \highlight{yellow}{44.69}/ 45.17 & \highlight{gray!30}{71.37}/ 70.25 \\
\cmidrule(lr){1-6}
Gold & \highlight{gray!30}{20.60}/ \highlight{gray!30}{23.40} & \highlight{yellow}{80.67}/ \highlight{yellow}{82.13} & \highlight{gray!30}{70.03}/ \highlight{yellow}{72.86} & 40.43 / 44.57 & \highlight{yellow}{69.67}/ \highlight{yellow}{72.66} \\ 
\midrule
\multicolumn{6}{c}{\textbf{Reddit ( 4B / 8B)}} \\
\midrule
NOSFT & 6.39 / 7.31 & 56.43 / 57.54 & 38.15 / 39.47 & 19.83 / 20.13 & 37.64 / 38.88 \\
\cmidrule(lr){1-6}
0 & 9.94 / 7.70 & 61.18 / 56.10 & 45.26 / 36.33 & 23.89 / 19.42 & 44.77 / 35.89 \\
20\% & 11.90 / 15.21 & 66.49 / 69.90 & 52.02 / 57.22 & 25.50 / 30.40 & 51.58 / 56.80 \\
40\% & 12.51 / 15.43 & 67.25 / 70.52 & 52.92 / 58.15 & 26.53 / 30.82 & 52.44 / 57.73 \\
60\% & 11.66 / 15.15 & 66.76 / 70.10 & 52.47 / 57.57 & 25.94 / 30.03 & 51.97 / 57.11 \\
80\% & 12.59 / \highlight{yellow}{17.20} & 68.48 / \highlight{gray!30}{72.40} & 55.14 / \highlight{gray!30}{61.26} & \highlight{gray!30}{27.25}/ \highlight{yellow}{32.81} & 54.65 / \highlight{yellow}{60.89} \\
100\% & 10.58 / 16.01 & 64.05 / 70.72 & 48.49 / 58.28 & 24.03 / 31.37 & 47.97 / 57.86 \\
120\% & \highlight{yellow}{14.57}/ \highlight{gray!30}{17.03} & \highlight{gray!30}{71.98}/ 71.58 & \highlight{yellow}{59.32}/ 59.69 & \highlight{yellow}{30.34}/ \highlight{gray!30}{32.13} & \highlight{yellow}{58.87}/ 59.32 \\
\cmidrule(lr){1-6}
Gold & \highlight{gray!30}{13.64}/ 16.39 & \highlight{yellow}{72.15}/ \highlight{yellow}{74.00} & \highlight{gray!30}{57.26}/ \highlight{yellow}{60.56} & 25.66 / 29.94 & \highlight{gray!30}{56.84}/ \highlight{gray!30}{60.22} \\
\bottomrule
\end{tabular}
\label{tab:semantic_metrics}
\end{table}

\section{Threats to Validity}
\subsection{Internal Validity}
Variability in quantity, quality, hallucinations, and residual noise of D-SCoRE-generated QA-CoT pairs across configurations may confound performance differences, especially favoring high-implicit or heterogeneous setups.
These threats were mitigated by fixed random seeds, consistent generation and training hyperparameters, uniform hardware, fixed-size random subsets (22,446 pairs), and rigorous Stage-2 quality control with regeneration. 
\subsection{External Validity}
Evaluation is limited to datasets with gold QA annotations and Qwen models, restricting generalizability to unannotated corpora or other LLM families (e.g., Llama, Mistral). Observed reasoning transfer and heterogeneous benefits may not hold across architectures or domain shifts.
Future validation on diverse models and annotation-free sources would strengthen external validity.
\subsection{Construct Validity}
F1 and EM metrics focus on token overlap and may undervalue semantic equivalence, reasoning depth, or hallucination resistance in implicit QA-CoT data. Gold annotations may introduce biases amplified by synthetic data.
We supplemented with BLEU, ROUGE, and semantic similarity metrics, confirming advantages for high-implicit configurations. Future work will incorporate human evaluation to better assess factuality, coherence, and reasoning quality.
\subsection{Conclusion Validity}
Randomness in LLMs generation (e.g., sampling temperature) and fine-tuning initialization may introduce variability, especially for small performance differences in high-implicit regimes. Fixed training budgets and random subset sampling may also affect reproducibility.
These issues were mitigated by consistent hyperparameters, multi-domain evaluation across model scales, and observation of clear monotonic trends (Figure~\ref{fig:experiment result}). Limitations include reliance on the Qwen model family and extractive-style benchmarks. Multiple independent runs with varied seeds, together with broader model and task diversity, would further strengthen statistical reliability and generalizability.
\section{Conclusion}
D-SCoRE revolutionizes domain QA dataset generation by producing implicit and explicit QA-CoT pairs from arbitrary texts, enabling efficient SFT. Via reasoning transfer, it outperforms human-annotated SQuAD Gold across domains using only one-third the data, while generating over 1,100 high-quality QA pairs per GPU-hour end-to-end on consumer-grade hardware. This highlights its superior efficiency, scalability, and the necessity of heterogeneous quality control to minimize synthetic noise.

Our experiments show D-SCoRE’s ability to produce effective QA pairs, with performance varying by dataset complexity and question ratios. Future work exploring refined ratio impacts and new domains could further enhance D-SCoRE’s robustness and generalizability, advancing automated domain QA generation and SFT for broader applications.

\clearpage
\bibliography{aaai2026}

%%%%%%%%%%%%%%%%%%%%%%%%%%%%%%%%%%%%%%%%%%%%%%%%%%%%%%%%%%%%%%%%%%%%%%%%%%%%%%%
%%%%%%%%%%%%%%%%%%%%%%%%%%%%%%%%%%%%%%%%%%%%%%%%%%%%%%%%%%%%%%%%%%%%%%%%%%%%%%%
% APPENDIX
%%%%%%%%%%%%%%%%%%%%%%%%%%%%%%%%%%%%%%%%%%%%%%%%%%%%%%%%%%%%%%%%%%%%%%%%%%%%%%%
%%%%%%%%%%%%%%%%%%%%%%%%%%%%%%%%%%%%%%%%%%%%%%%%%%%%%%%%%%%%%%%%%%%%%%%%%%%%%%%
\newpage
\appendix
\onecolumn
\section{Appendix}
\subsection{Complete Experiment Results}
\begin{table}[h!]
\centering
\small
\caption{Performance improvement of heterogeneous over homogeneous pipelines ( $ \Delta = \text{heterogeneous} - \text{homogeneous} $ ) across implicit ratios and model scales for multiple evaluation metrics. Within each metric and model size, the highest and second-highest values are highlighted in yellow and gray, respectively.}
\label{tab:full_delta}
\begin{tabular}{@{} c 
    @{\hspace{12pt}} c @{\hspace{6pt}} c 
    @{\hspace{12pt}} c @{\hspace{6pt}} c 
    @{\hspace{12pt}} c @{\hspace{6pt}} c 
    @{\hspace{12pt}} c @{\hspace{6pt}} c 
    @{\hspace{12pt}} c @{\hspace{6pt}} c 
    @{\hspace{12pt}} c @{\hspace{6pt}} c 
    @{\hspace{12pt}} c @{\hspace{6pt}} c 
    @{}}
\toprule
\textbf{Imp.\%} &
\multicolumn{2}{c}{\textbf{$\Delta$ F1}} &
\multicolumn{2}{c}{\textbf{$\Delta$ EM}} &
\multicolumn{2}{c}{\textbf{$\Delta$ BLEU}} &
\multicolumn{2}{c}{\textbf{$\Delta$ SemSim}} &
\multicolumn{2}{c}{\textbf{$\Delta$ ROUGE-1}} &
\multicolumn{2}{c}{\textbf{$\Delta$ ROUGE-2}} &
\multicolumn{2}{c}{\textbf{$\Delta$ ROUGE-L}} \\
\cmidrule(lr){2-3} \cmidrule(lr){4-5} \cmidrule(lr){6-7} \cmidrule(lr){8-9} \cmidrule(lr){10-11} \cmidrule(lr){12-13} \cmidrule(lr){14-15}
& 4B & 8B & 4B & 8B & 4B & 8B & 4B & 8B & 4B & 8B & 4B & 8B & 4B & 8B \\
\midrule
\multicolumn{15}{c}{\textbf{Amazon}} \\
\cmidrule(lr){1-15}
0\%   & 7.11 & 5.48 & 3.58 & 3.72 & 5.54 & 3.97 & 5.59 & -2.59 & -1.54 & 0.65 & -1.67 & -3.29 & -0.26 & -3.19 \\
20\%  & 14.89 & 10.38 & 5.98 & 6.58 & 14.33 & 8.77 & 14.36 & 4.94 & 5.40 &  \highlight{gray!30}{5.00} & 2.51 & 4.53 & 3.88 & 4.67 \\
40\%  & 8.53 & 7.35 & 4.51 & 4.66 & 8.41 & 6.49 & 8.42 & 1.33 & 0.43 & 2.31 & -0.15 & 0.77 & 1.65 & 0.79 \\
60\%  & 12.70 & 13.01 & 5.16 & 8.40 & 12.53 & 7.29 & 12.62 & 1.62 & 1.80 & 2.74 & -0.10 & 1.15 & 1.11 & 1.26 \\
80\%  &  \highlight{gray!30}{23.06} &  \highlight{gray!30}{22.44} &  \highlight{gray!30}{7.87} &  \highlight{gray!30}{14.12} &  \highlight{gray!30}{23.44} &  \highlight{gray!30}{12.72} &  \highlight{gray!30}{23.53} & \highlight{yellow}{8.00} &  \highlight{gray!30}{7.63} & 4.81 & \highlight{yellow}{5.08} & \highlight{yellow}{8.09} & \highlight{yellow}{5.92} & \highlight{yellow}{8.19} \\
100\% & 13.47 & 11.21 & 5.21 & 7.30 & 13.21 & 7.86 & 13.28 & 4.17 & 3.74 & 3.92 & 2.59 & 4.07 & 2.93 & 4.10 \\
120\% & \highlight{yellow}{28.59} & \highlight{yellow}{28.43} & \highlight{yellow}{9.77} & \highlight{yellow}{18.38} & \highlight{yellow}{29.08} & \highlight{yellow}{15.89} & \highlight{yellow}{29.14} &  \highlight{gray!30}{7.16} &  \highlight{yellow}{7.71} & \highlight{yellow}{5.33} &  \highlight{gray!30}{4.38} &  \highlight{gray!30}{6.73} &  \highlight{gray!30}{5.51} &  \highlight{gray!30}{6.87} \\
\midrule
\multicolumn{15}{c}{\textbf{New Wiki}} \\
\cmidrule(lr){1-15}
0\%   & 7.26 & 6.45 & 5.60 & 4.51 & 5.78 & 4.06 & 5.84 & -0.91 & 1.06 & 2.22 & 0.02 & -1.78 & -0.25 & -1.63 \\
20\%  & 16.69 & 14.36 & 8.95 & 7.93 & 16.31 & 11.24 & 16.44 & \highlight{gray!30}{7.32} & \highlight{gray!30}{8.69} & \highlight{gray!30}{6.62} & \highlight{gray!30}{4.50} & \highlight{gray!30}{6.92} & \highlight{gray!30}{5.87} & \highlight{gray!30}{7.12} \\
40\%  & 8.51 & 8.54 & 5.36 & 5.20 & 8.55 & 5.36 & 8.56 & 2.24 & 1.41 & 2.48 & 0.83 & 1.73 & 1.94 & 1.74 \\
60\%  & 14.16 & 15.75 & 7.22 & 9.14 & 14.57 & 9.21 & 14.69 & 3.88 & 4.03 & 4.51 & 2.08 & 3.56 & 3.68 & 3.72 \\
80\%  &  \highlight{gray!30}{22.77} & \highlight{gray!30}{23.13} & \highlight{gray!30}{10.06} & \highlight{gray!30}{13.46} & \highlight{gray!30}{23.28} & \highlight{gray!30}{14.97} & \highlight{gray!30}{23.36} & \highlight{yellow}{8.70} & \highlight{yellow}{9.65} & \highlight{yellow}{6.80} & \highlight{yellow}{5.68} & \highlight{yellow}{8.48} & \highlight{yellow}{6.34} & \highlight{yellow}{8.70} \\
100\% & 11.51 & 9.52 & 6.19 & 6.10 & 11.39 & 8.78 & 11.44 & 4.30 & 4.69 & 4.62 & 3.02 & 4.20 & 3.56 & 4.34 \\
120\% & \highlight{yellow}{28.07} & \highlight{yellow}{29.73} & \highlight{yellow}{13.26} & \highlight{yellow}{17.62} & \highlight{yellow}{28.41} & \highlight{yellow}{18.89} & \highlight{yellow}{28.66} & 6.87 & 7.56 & 6.08 & 4.24 & 6.49 & 5.16 & 6.69 \\
\midrule
\multicolumn{15}{c}{\textbf{NYT}} \\
\cmidrule(lr){1-15}
0\%   & 8.56 & 8.64 & 5.27 & 5.98 & 7.05 & 5.43 & 7.12 & -6.39 & -3.44 & -0.37 & -2.47 & -6.70 & -3.02 & -6.59 \\
20\%  & 15.51 & 14.11 & 7.39 & 8.94 & 15.74 & 10.27 & 15.79 & 5.68 & 6.70 & \highlight{yellow}{5.22} & 3.34 & 5.23 & \highlight{gray!30}{4.46} & 5.36 \\
40\%  & 6.28 & 6.14 & 4.07 & 4.18 & 6.25 & 4.25 & 6.32 & 0.97 & -0.38 & 1.80 & 0.16 & 0.52 & 1.06 & 0.52 \\
60\%  & 11.95 & 13.02 & 5.78 & 8.49 & 12.30 & 7.66 & 12.34 & 1.78 & 1.93 & 3.02 & 0.92 & 1.79 & 2.05 & 1.82 \\
80\%  & \highlight{gray!30}{22.19} & \highlight{gray!30}{23.81} & \highlight{gray!30}{9.11} & \highlight{gray!30}{14.86} & \highlight{gray!30}{22.88} & \highlight{gray!30}{14.45} & \highlight{gray!30}{22.96} & \highlight{yellow}{7.57} & \highlight{yellow}{8.37} & 4.85 & \highlight{yellow}{5.56} & \highlight{yellow}{7.45} & \highlight{yellow}{5.60} & \highlight{yellow}{7.51} \\
100\% & 10.64 & 9.78 & 5.07 & 6.50 & 10.78 & 7.35 & 10.81 & 4.49 & 4.67 & 4.03 & 3.40 & 4.48 & 3.33 & 4.51 \\
120\% & \highlight{yellow}{27.54} & \highlight{yellow}{30.05} & \highlight{yellow}{11.70} & \highlight{yellow}{18.85} & \highlight{yellow}{28.17} & \highlight{yellow}{18.20} & \highlight{yellow}{28.35} & \highlight{gray!30}{5.72} & \highlight{gray!30}{7.00} & \highlight{gray!30}{4.99} & \highlight{gray!30}{3.79} & \highlight{gray!30}{5.48} & 4.45 & \highlight{gray!30}{5.53} \\
\midrule
\multicolumn{15}{c}{\textbf{Reddit}} \\
\cmidrule(lr){1-15}
0\%   & 7.08 & 4.51 & 2.97 & 3.95 & 5.64 & 4.10 & 5.71 & -3.22 & -3.69 & -0.06 & -1.68 & -3.88 & -0.70 & -3.77 \\
20\%  & 12.20 & 10.90 & 4.92 & 6.92 & 11.87 & 6.56 & 11.95 & 5.76 & 5.82 & 4.05 & 3.42 & 5.13 & 4.41 & 5.23 \\
40\%  & 7.28 & 6.43 & 3.58 & 4.74 & 7.12 & 4.52 & 7.16 & 2.10 & 1.45 & 2.12 & 1.22 & 1.74 & 2.00 & 1.76 \\
60\%  & 10.52 & 10.16 & 4.08 & 7.11 & 10.89 & 6.63 & 10.93 & 2.64 & 2.80 & 2.41 & 1.27 & 2.21 & 1.30 & 2.18 \\
80\%  &  \highlight{gray!30}{18.81} &  \highlight{gray!30}{19.93} &  \highlight{gray!30}{6.48} &  \highlight{gray!30}{12.22} &  \highlight{gray!30}{19.08} &  \highlight{gray!30}{10.24} &  \highlight{gray!30}{19.20} & \highlight{yellow}{9.03} & \highlight{yellow}{9.39} & \highlight{yellow}{5.26} & \highlight{yellow}{5.96} & \highlight{yellow}{8.86} & \highlight{yellow}{6.28} & \highlight{yellow}{8.98} \\
100\% & 11.63 & 10.49 & 4.40 & 7.26 & 11.60 & 7.00 & 11.67 & 5.14 & 5.61 & 4.22 & 3.46 & 4.90 & 4.22 & 5.05 \\
120\% & \highlight{yellow}{25.45} & \highlight{yellow}{26.77} & \highlight{yellow}{9.33} & \highlight{yellow}{17.54} & \highlight{yellow}{26.05} & \highlight{yellow}{14.80} & \highlight{yellow}{26.19} &  \highlight{gray!30}{7.29} &  \highlight{gray!30}{7.65} &  \highlight{gray!30}{5.07} &  \highlight{gray!30}{4.92} &  \highlight{gray!30}{7.18} &  \highlight{gray!30}{5.49} &  \highlight{gray!30}{7.35} \\
\midrule
\multicolumn{15}{c}{\textbf{SQuAD}} \\
\cmidrule(lr){1-15}
0\%   & 9.09 & 10.47 & 6.02 & 6.40 & 7.55 & 4.67 & 7.60 & -2.39 & -0.44 & 1.19 & -0.43 & -3.13 & -1.44 & -3.04 \\
20\%  & 16.55 & 15.85 & 8.61 & 8.90 & 16.33 & 11.12 & 16.37 & 6.68 & 8.27 & \highlight{gray!30}{5.54} & \highlight{gray!30}{4.86} & 6.55 & \highlight{gray!30}{5.23} & 6.60 \\
40\%  & 9.19 & 10.95 & 5.48 & 6.21 & 8.97 & 6.18 & 9.02 & 1.00 & 0.77 & 1.71 & 0.28 & 0.66 & 1.20 & 0.75 \\
60\%  & 15.88 & 19.15 & 7.92 & 11.02 & 16.11 & 9.89 & 16.15 & 3.91 & 4.69 & 3.61 & 2.41 & 3.77 & 2.74 & 3.86 \\
80\%  & \highlight{gray!30}{26.11} & \highlight{gray!30}{29.05} & \highlight{gray!30}{11.32} & \highlight{gray!30}{16.98} & \highlight{gray!30}{26.54} & \highlight{gray!30}{17.21} & \highlight{gray!30}{26.60} & \highlight{yellow}{11.27} & \highlight{yellow}{13.78} & \highlight{yellow}{6.77} & \highlight{yellow}{8.29} & \highlight{yellow}{11.11} & \highlight{yellow}{7.40} & \highlight{yellow}{11.17} \\
100\% & 10.64 & 10.27 & 5.57 & 5.84 & 10.62 & 8.12 & 10.64 & 5.02 & 6.27 & 4.35 & 3.75 & 5.05 & 3.43 & 5.06 \\
120\% & \highlight{yellow}{32.01} & \highlight{yellow}{37.06} & \highlight{yellow}{14.00} & \highlight{yellow}{22.01} & \highlight{yellow}{32.31} & \highlight{yellow}{21.03} & \highlight{yellow}{32.36} & \highlight{gray!30}{6.76} & \highlight{gray!30}{8.55} & 5.08 & \highlight{gray!30}{4.85} & 6.76 & 4.62 & \highlight{gray!30}{6.83} \\
\bottomrule
\end{tabular}
\end{table}

\newpage

\begin{table}[H]
\centering
\small
\caption{Performance of 4B and 8B models across datasets, implicit ratios, and evaluation metrics. Each cell reports results as 4B / 8B. Within each dataset and metric, the highest and second-highest values are independently identified for the 4B and 8B models across all implicit ratio configurations, and highlighted in yellow and gray, respectively.}
\label{tab:full_metrics}
\begin{tabular}{cccccccc}
\toprule
% Config & F1  $ \uparrow $  & EM  $ \uparrow $  & BLEU  $ \uparrow $  & SemSim  $ \uparrow $  & ROUGE-1  $ \uparrow $  & ROUGE-2  $ \uparrow $  & ROUGE-L  $ \uparrow $  \\
\textbf{Config} & \textbf{F1 $\uparrow$} & \textbf{EM  $\uparrow$} & \textbf{BLEU $\uparrow$} & \textbf{SemSim $\uparrow$} & \textbf{ROUGE-1 $\uparrow$} & \textbf{ROUGE-2 $\uparrow$} & \textbf{ROUGE-L $\uparrow$} \\
\midrule
\multicolumn{8}{c}{\textbf{Amazon (4B / 8B)}} \\
\midrule
NOSFT & 42.16 / 42.27 & 11.27 / 13.17 & 7.63 / 8.74 & 56.97 / 58.11 & 41.04 / 41.86 & 23.32 / 23.51 & 40.53 / 41.31 \\
\cmidrule(lr){1-8}
0\%   & 49.83 / 38.42 & 22.04 / 12.36 & 11.90 / 8.82 & 62.84 / 56.08 & 49.05 / 37.85 & 27.32 / 22.06 & 48.63 / 37.41 \\
20\%  & 52.42 / 57.74 & 30.05 / 35.21 & 12.62 / 16.47 & 66.68 / 69.78 & 52.68 / 58.34 & 27.20 / 32.71 & 52.26 / 57.93 \\
40\%  & 54.10 / 57.42 & 33.14 / 34.42 & 13.91 / 16.23 & 68.27 / 69.43 & 54.78 / 57.86 & 28.94 / 32.33 & 54.32 / 57.44 \\
60\%  & 54.30 / 57.74 & 33.61 / 35.37 & 13.22 / 16.40 & 68.67 / 69.69 & 55.00 / 58.23 & 28.28 / 32.19 & 54.53 / 57.84 \\
80\%  & \highlight{gray!30}{56.02} / \highlight{yellow}{60.71} & \highlight{gray!30}{35.67} / 37.43 & 13.64 / 17.78 & 69.54 / \highlight{gray!30}{71.84} & 56.93 / \highlight{gray!30}{61.39} & \highlight{gray!30}{29.04} / \highlight{yellow}{34.73} & 56.46 / \highlight{gray!30}{61.05} \\
100\% & 48.70 / 58.04 & 26.55 / 34.26 & 11.67 / 16.77 & 64.09 / 70.21 & 49.06 / 58.80 & 25.80 / 32.87 & 48.54 / 58.35 \\
120\% & \highlight{yellow}{59.05} / \highlight{gray!30}{59.98} & \highlight{yellow}{38.18} / \highlight{gray!30}{37.52} & \highlight{gray!30}{15.04} / \highlight{gray!30}{18.10} & \highlight{gray!30}{71.87} / 71.50 & \highlight{yellow}{60.01} / 60.50 & \highlight{yellow}{31.18} / \highlight{gray!30}{34.43} & \highlight{yellow}{59.54} / 60.16 \\
\cmidrule(lr){1-8}
Gold  & 54.79 / 59.58 & 33.49 / \highlight{yellow}{38.37} & \highlight{yellow}{15.13} / \highlight{yellow}{18.50} & \highlight{yellow}{74.50} / \highlight{yellow}{76.19} & \highlight{gray!30}{58.90} / \highlight{yellow}{63.25} & 27.79 / 32.54 & \highlight{gray!30}{58.50} / \highlight{yellow}{63.03} \\
\midrule
\multicolumn{8}{c}{\textbf{New Wiki (4B / 8B)}} \\
\midrule
NOSFT & 51.37 / 50.18 & 21.54 / 22.70 & 12.79 / 14.67 & 64.91 / 64.60 & 50.51 / 50.31 & 34.31 / 33.61 & 50.05 / 49.74 \\
\cmidrule(lr){1-8}
0\%   & 51.17 / 49.04 & 24.94 / 22.17 & 13.31 / 13.32 & 65.56 / 64.00 & 52.26 / 49.76 & 34.90 / 32.66 & 51.78 / 49.13 \\
20\%  & 46.54 / 58.51 & 27.41 / 35.10 & 11.13 / 17.03 & 66.68 / 71.39 & 47.34 / 59.44 & 29.41 / 38.03 & 46.65 / 58.80 \\
40\%  & \highlight{gray!30}{55.33} / \highlight{gray!30}{63.49} & \highlight{gray!30}{34.52} / \highlight{gray!30}{41.72} & \highlight{gray!30}{15.31} / \highlight{gray!30}{20.53} & \highlight{gray!30}{70.25} / \highlight{gray!30}{75.12} & \highlight{gray!30}{55.94} / \highlight{gray!30}{64.47} & \highlight{gray!30}{35.58} / \highlight{gray!30}{41.43} & \highlight{gray!30}{55.34} / \highlight{gray!30}{63.99} \\
60\%  & 50.14 / 62.37 & 27.34 / 40.19 & 12.94 / 19.16 & 66.85 / 74.53 & 50.81 / 63.29 & 32.38 / 40.53 & 50.13 / 62.69 \\
80\%  & 41.98 / 59.15 & 20.56 / 35.85 & 10.28 / 18.25 & 62.72 / 71.88 & 42.56 / 59.97 & 26.71 / 38.90 & 41.92 / 59.37 \\
100\% & 45.91 / 61.23 & 24.44 / 37.84 & 11.58 / 18.97 & 64.58 / 73.11 & 46.57 / 61.88 & 29.33 / 40.22 & 45.85 / 61.32 \\
120\% & 39.51 / 60.32 & 17.08 / 37.52 & 9.27 / 18.61 & 60.81 / 72.84 & 40.16 / 61.21 & 24.86 / 39.49 & 39.42 / 60.64 \\
\cmidrule(lr){1-8}
Gold  & \highlight{yellow}{65.65} / \highlight{yellow}{68.96} & \highlight{yellow}{45.12} / \highlight{yellow}{48.12} & \highlight{yellow}{21.51} / \highlight{yellow}{24.97} & \highlight{yellow}{79.41} / \highlight{yellow}{81.03} & \highlight{yellow}{68.04} / \highlight{yellow}{70.59} & \highlight{yellow}{38.96} / \highlight{yellow}{42.95} & \highlight{yellow}{67.59} / \highlight{yellow}{70.24} \\
\midrule
\multicolumn{8}{c}{\textbf{NYT (4B / 8B)}} \\
\midrule
NOSFT & 52.15 / 50.37 & 24.44 / 24.70 & 11.48 / 12.89 & 65.66 / 64.23 & 51.80 / 50.82 & 33.79 / 32.71 & 51.52 / 50.50 \\
\cmidrule(lr){1-8}
0\%   & 60.32 / 45.20 & 36.23 / 23.43 & 17.52 / 12.40 & 71.89 / 62.37 & 60.81 / 45.71 & 39.98 / 30.12 & 60.57 / 45.45 \\
20\%  & 65.73 / 69.13 & 46.66 / 49.76 & 19.36 / 22.57 & 77.07 / 78.45 & 67.02 / 70.22 & 41.83 / \highlight{gray!30}{45.35} & 66.72 / 69.96 \\
40\%  & 65.78 / 68.75 & 47.14 / 48.55 & 19.83 / 22.18 & 77.45 / 78.36 & 67.18 / 69.87 & 42.03 / 44.93 & 66.90 / 69.61 \\
60\%  & 67.63 / 68.81 & 49.01 / 49.23 & 19.94 / 22.35 & 78.67 / 78.52 & 69.11 / 69.99 & 42.72 / 45.02 & 68.80 / 69.70 \\
80\%  & \highlight{gray!30}{67.99} / \highlight{yellow}{70.76} & \highlight{gray!30}{49.78} / \highlight{gray!30}{51.01} & 20.16 / \highlight{gray!30}{23.40} & 78.95 / \highlight{gray!30}{79.95} & 69.67 / \highlight{gray!30}{72.01} & \highlight{gray!30}{43.36} / \highlight{yellow}{46.43} & 69.39 / \highlight{gray!30}{71.74} \\
100\% & 61.32 / 68.66 & 41.24 / 48.13 & 17.64 / 22.60 & 73.50 / 78.50 & 62.64 / 70.01 & 39.34 / 45.06 & 62.31 / 69.71 \\
120\% & \highlight{yellow}{70.01} / 69.29 & \highlight{yellow}{52.54} / 50.02 & \highlight{yellow}{21.26} / 23.32 & \highlight{gray!30}{80.44} / 78.82 & \highlight{yellow}{71.64} / 70.48 & \highlight{yellow}{44.69} / 45.17 & \highlight{yellow}{71.37} / 70.25 \\
\cmidrule(lr){1-8}
Gold  & 66.79 / \highlight{gray!30}{70.46} & 48.15 / \highlight{yellow}{52.35} & \highlight{gray!30}{20.60} / \highlight{yellow}{23.40} & \highlight{yellow}{80.67} / \highlight{yellow}{82.13} & \highlight{gray!30}{70.03} / \highlight{yellow}{72.86} & 40.43 / 44.57 & \highlight{gray!30}{69.67} / \highlight{yellow}{72.66} \\
\midrule
\multicolumn{8}{c}{\textbf{Reddit (4B / 8B)}} \\
\midrule
NOSFT & 38.43 / 39.23 & 10.43 / 13.81 & 6.39 / 7.31 & 56.43 / 57.54 & 38.15 / 39.47 & 19.83 / 20.13 & 37.64 / 38.88 \\
\cmidrule(lr){1-8}
0\%   & 45.56 / 36.31 & 19.32 / 12.24 & 9.94 / 7.70 & 61.18 / 56.10 & 45.26 / 36.33 & 23.89 / 19.42 & 44.77 / 35.89 \\
20\%  & 51.24 / 56.43 & 30.40 / 35.87 & 11.90 / 15.21 & 66.49 / 69.90 & 52.02 / 57.22 & 25.50 / 30.40 & 51.58 / 56.80 \\
40\%  & 51.85 / 57.32 & 30.65 / 36.30 & 12.51 / 15.43 & 67.25 / 70.52 & 52.92 / 58.15 & 26.53 / 30.82 & 52.44 / 57.73 \\
60\%  & 51.19 / 56.77 & 30.28 / 35.58 & 11.66 / 15.15 & 66.76 / 70.10 & 52.47 / 57.57 & 25.94 / 30.03 & 51.97 / 57.11 \\
80\%  & \highlight{gray!30}{53.99} / \highlight{yellow}{60.41} & 33.61 / \highlight{yellow}{39.38} & 12.59 / \highlight{yellow}{17.20} & 68.48 / \highlight{gray!30}{72.40} & 55.14 / \highlight{yellow}{61.26} & \highlight{gray!30}{27.25} / \highlight{yellow}{32.81} & 54.65 / \highlight{yellow}{60.89} \\
100\% & 47.73 / 57.44 & 25.48 / 35.91 & 10.58 / 16.01 & 64.05 / 70.72 & 48.49 / 58.28 & 24.03 / 31.37 & 47.97 / 57.86 \\
\cmidrule(lr){1-8}
120\% & \highlight{yellow}{57.95} / \highlight{gray!30}{58.73} & \highlight{yellow}{37.07} / 37.73 & \highlight{yellow}{14.57} / \highlight{gray!30}{17.03} & \highlight{gray!30}{71.98} / 71.58 & \highlight{yellow}{59.32} / 59.69 & \highlight{yellow}{30.34} / \highlight{gray!30}{32.13} & \highlight{yellow}{58.87} / 59.32 \\
Gold  & 53.61 / 57.76 & \highlight{gray!30}{34.18} / \highlight{gray!30}{38.20} & \highlight{gray!30}{13.64} / 16.39 & \highlight{yellow}{72.15} / \highlight{yellow}{74.00} & \highlight{gray!30}{57.26} / \highlight{gray!30}{60.56} & 25.66 / 29.94 & \highlight{gray!30}{56.84} / \highlight{gray!30}{60.22} \\
\midrule
\multicolumn{8}{c}{\textbf{SQuAD (4B / 8B)}} \\
\midrule
NOSFT & 55.22 / 52.15 & 28.67 / 28.59 & 12.04 / 14.12 & 68.09 / 65.10 & 53.71 / 51.82 & 35.27 / 34.07 & 53.53 / 51.64 \\
0\%   & 65.13 / 50.22 & 44.25 / 29.60 & 19.10 / 14.57 & 75.14 / 64.87 & 64.51 / 49.76 & 41.82 / 32.68 & 64.36 / 49.63 \\
\cmidrule(lr){1-8}
20\%  & 72.81 / 72.79 & 56.81 / 56.09 & 22.72 / 24.68 & 81.48 / 80.57 & 72.79 / 72.87 & 46.65 / 47.66 & 72.57 / 72.68 \\
40\%  & 74.14 / 73.36 & 58.99 / 56.52 & 23.79 / 24.68 & 82.74 / 81.05 & 74.06 / 73.35 & 47.73 / 47.91 & 73.84 / 73.19 \\
60\%  & 75.18 / 74.21 & 60.07 / 57.64 & 23.44 / 25.11 & 83.40 / 81.62 & 75.36 / 74.31 & 47.73 / 48.40 & 75.16 / 74.15 \\
80\%  & 75.41 / \highlight{gray!30}{76.25} & \highlight{gray!30}{60.67} / \highlight{gray!30}{59.85} & 23.63 / \highlight{yellow}{26.46} & 83.60 / \highlight{gray!30}{83.14} & 75.73 / \highlight{gray!30}{76.35} & \highlight{gray!30}{48.39} / \highlight{yellow}{49.61} & 75.53 / \highlight{gray!30}{76.17} \\
100\% & 64.90 / 72.86 & 47.25 / 55.54 & 19.79 / 24.81 & 75.31 / 80.71 & 64.80 / 72.99 & 42.52 / 47.78 & 64.56 / 72.80 \\
120\% & \highlight{yellow}{77.97} / 74.88 & \highlight{yellow}{64.14} / 58.73 & \highlight{yellow}{25.22} / 25.92 & \highlight{gray!30}{85.68} / 82.32 & \highlight{yellow}{78.17} / 75.05 & \highlight{yellow}{49.89} / \highlight{gray!30}{48.92} & \highlight{yellow}{77.94} / 74.88 \\
\cmidrule(lr){1-8}
Gold  & \highlight{gray!30}{76.25} / \highlight{yellow}{77.70} & 60.34 / \highlight{yellow}{62.14} & \highlight{gray!30}{24.63} / \highlight{gray!30}{26.01} & \highlight{yellow}{86.34} / \highlight{yellow}{86.69} & \highlight{gray!30}{78.00} / \highlight{yellow}{78.98} & 46.27 / 47.59 & \highlight{gray!30}{77.65} /  \highlight{yellow}{78.82} \\
\bottomrule
\end{tabular}
\label{tab:semantic_and_task_metrics}
\end{table}

\newpage

\subsection{Prompt Templates}

\subsubsection{QA Pair Generation Prompt}

D-SCoRE generates high-quality question-answer pairs using the following structured prompt, which enforces strict guidelines for entity selection, question clarity, and reasoning depth.

\begin{tcblisting}{
    breakable,
    enhanced,
    boxrule=0.4pt,
    colback=gray!6,
    colframe=gray!40,
    arc=3pt,
    fontupper=\ttfamily\small,
    listing only,
    title={Structured QA Generation (2 Explicit + 1 Implicit)},
    listing options={numbers=none}
}
You are a QA generation expert specialized in creating high-quality question-answer pairs.
Generate exactly 3 QA pairs: 2 explicit and 1 implicit.

**CRITICAL**: All questions must be **CLEAR AND UNAMBIGUOUS** - they should contain sufficient context to be understood without confusion.

**Content:**
\{content\}

**Requirements:**

**Explicit QA (2 pairs):**
- **Entity Selection Strategy**: Choose entities that will be the **PRIMARY ANSWER** to your question:
  * **Core Principle**: The entity should be the most specific and informative answer possible
  * **Specificity Priority**: Choose the most specific entity available
  * **Answer-Focused**: The entity should be what you want the model to learn to extract/identify
  
- **Question Design**: Create questions that:
  * **Target the entity as the primary answer**
  * Include sufficient context and specific identifiers
  * Avoid vague pronouns or references
  * **MUST align with the entity and expected answer type**
  
- **Answer Requirements**: 
  * **CRITICAL**: Answer should directly correspond to the selected entity
  * **Specificity Rule**: Use the most specific answer available
  * **Entity-Answer Consistency**: The answer should match or include the entity
  * **Completeness**: Ensure answers are complete and precise

**Implicit QA (1 pair):**  
- **Content Check**: Ensure sufficient information exists for genuine logical inference
- **Question Types**: Focus on:
  * Causal relationships, comparative analysis, inferential conclusions
  * Logical deductions from multiple facts
- **Answer Quality**: Derived through reasoning, not directly stated

**Paraphrase Guidelines:**
- **Semantic Role Shift**: Transform the question by shifting focus while maintaining the same answer
- **Preserve Context**: Keep all necessary contextual information and specificity  
- **Answer Consistency**: **CRITICAL** - paraphrased question must yield identical answer
- **Natural Language**: Use fluent, natural phrasing
- **Quality Check**: Maximum 40\% word overlap, prioritize semantic accuracy

**Output Format (JSON):**
```json
[
  \{
    "type": "explicit",
    "attributes": \{
      "entity": "Most specific entity that is the primary answer"
    \},
    "question": "Your question targeting the entity as primary answer?",
    "paraphrases": "Your paraphrased question with identical answer?",
    "a0": "Answer that matches or includes the entity"
  \},
  \{
    "type": "explicit", 
    "attributes": \{
      "entity": "Second most specific entity as primary answer"
    \},
    "question": "Your second question targeting the entity?",
    "paraphrases": "Your second paraphrased question with identical answer?",
    "a0": "Second answer that matches or includes the entity"
  \},
  \{
    "type": "implicit",
    "attributes": \{
      "reasoning\_trace": "Step 1: [Identify relevant facts] Step 2: [Apply logical reasoning] Step 3: [Draw conclusion] Therefore: [Final inference]"
    \},
    "question": "Your reasoning question with sufficient context?",
    "paraphrases": "Your paraphrased reasoning question with same conclusion?",
    "a0": "Answer derived through reasoning"
  \}
]

\end{tcblisting}

\begin{tcblisting}{
    breakable,
    enhanced,
    boxrule=0.4pt,
    colback=gray!6,
    colframe=gray!40,
    arc=3pt,
    fontupper=\ttfamily\small,
    listing only,
    title={Implicit-Only QA Generation (3 Pairs)},
    listing options={numbers=none}
}
You are a QA generation expert specialized in creating high-quality question-answer pairs.
Generate exactly 3 implicit QA pairs:

**CRITICAL**: All questions must be **CLEAR AND UNAMBIGUOUS** - they should contain sufficient context to be understood without confusion.

**Content:**
\{content\}

**Requirements:**

**Implicit QA (3 pairs):**  
- **Content Check**: Ensure sufficient information exists for genuine logical inference
- **Question Types**: Focus on:
  * Causal relationships, comparative analysis, inferential conclusions
  * Logical deductions from multiple facts
- **Answer Quality**: Derived through reasoning, not directly stated

**Paraphrase Guidelines:**
- **Semantic Role Shift**: Transform the question by shifting focus while maintaining the same answer
- **Preserve Context**: Keep all necessary contextual information and specificity  
- **Answer Consistency**: **CRITICAL** - paraphrased question must yield identical answer
- **Natural Language**: Use fluent, natural phrasing
- **Quality Check**: Maximum 40\% word overlap, prioritize semantic accuracy

**Output Format (JSON):**
```json
[
  \{
    "type": "implicit",
    "attributes": \{
      "reasoning\_trace": "Step 1: [Identify relevant facts] Step 2: [Apply logical reasoning] Step 3: [Draw conclusion] Therefore: [Final inferred answer]"
    \},
    "question": "Reasoning-based question with sufficient context?",
    "paraphrases": "Paraphrased question yielding the same answer?",
    "a0": "Answer derived through reasoning"
  \},
  \{
    "type": "implicit",
    "attributes": \{
      "reasoning\_trace": "Step 1: [Identify relevant facts] Step 2: [Apply logical reasoning] Step 3: [Draw conclusion] Therefore: [Final inferred answer]"
    \},
    "question": "Second reasoning-based question with sufficient context?",
    "paraphrases": "Second paraphrased question yielding the same answer?",
    "a0": "Second answer derived through reasoning"
  \},
  \{
    "type": "implicit",
    "attributes": \{
      "reasoning\_trace": "Step 1: [Identify relevant facts] Step 2: [Apply logical reasoning] Step 3: [Draw conclusion] Therefore: [Final inferred answer]"
    \},
    "question": "Third reasoning-based question with sufficient context?",
    "paraphrases": "Third paraphrased question yielding the same answer?",
    "a0": "Third answer derived through reasoning"
  \}
]
\end{tcblisting}

\begin{tcblisting}{
    breakable,
    enhanced,
    boxrule=0.4pt,
    colback=gray!6,
    colframe=gray!40,
    arc=3pt,
    fontupper=\ttfamily\small,
    listing only,
    title={Single Explicit QA Pair Generation},
    listing options={numbers=none}
}
Generate ONE high-quality explicit QA pair from the content below.

**Content:**
\{content\}

**Requirements:**
- Entity must be the primary answer target
- Answer must be directly extractable from content  
- Question must be clear and specific
- Use the most specific entity/answer available

**Output Format:**
```json
\{
"type": "explicit",
"attributes": \{"entity": "specific\_entity"\},
"question": "Your clear question?",
"paraphrases": "Paraphrased question with same answer?",
"a0": "Direct answer from content"
\}
\end{tcblisting}

\subsubsection{QA Quality Filtering Prompt}

D-SCoRE employs a strict quality control step using the following prompt to filter or correct generated QA pairs. The template uses \texttt{\{content\}} for the source passage and dynamically lists all QA pairs with their metadata.

\begin{tcblisting}{
    breakable,
    enhanced,
    boxrule=0.4pt,
    colback=gray!6,
    colframe=gray!40,
    arc=3pt,
    fontupper=\ttfamily\small,
    listing only,
    title={QA Filtering and Type Validation},
    listing options={numbers=none}
}
You are a QA quality assessor. Analyze each QA pair with STRICT standards.

**Content:**
\{content\}

**QA Pairs to Analyze:**
\{qa\_pairs\_formatted\}

**STRICT EVALUATION RULES:**

**For Explicit QA pairs:**
1. **Entity-Answer Consistency**: Entity must match the main answer component
2. **Direct Extraction**: Answer must be directly extractable from content (word-for-word or paraphrased)
3. **Question Quality**: Clear, specific, unambiguous
4. **Attributes**: Must have valid "entity" field

**For Implicit QA pairs:**
1. **TRUE INFERENCE REQUIREMENT**:
   - Must require combining multiple facts OR logical deduction OR causal reasoning
   - Must involve synthesis of information across different parts of text
2. **Reasoning Quality**: Must have clear, logical "reasoning\_trace"
3. **Complex Analysis**: Should involve interpretation, comparison, or conclusion-drawing

**CRITICAL TYPEFIX DETECTION:**

**Test for Fake Implicit (should be TYPEFIX to explicit):**
- Can you find the answer or its paraphrase directly in the text?
- Is the "reasoning" just restating what's already written?
- Does the question ask for information that's explicitly provided?

**Test for True Implicit (should remain implicit):**
- Does answering require combining info from multiple sentences?
- Must you make logical connections not explicitly stated?
- Does it require comparing, contrasting, or drawing conclusions?

**STRICT DECISION CRITERIA:**
- **KEEP**: QA meets all quality standards for its declared type
- **DELETE**: Critical flaws (wrong entity-answer match, poor quality, missing required fields)
- **TYPEFIX**: Wrong type classification AND new\_type differs from current type

**CRITICAL TYPEFIX RULES:**
- \item \textbf{Implicit $\rightarrow$ Explicit}: If answer can be directly found/paraphrased from content
- \item \textbf{Explicit $\rightarrow$ Implicit}: If answer requires true reasoning and cannot be directly extracted
- **TYPEFIX must change the type**: new\_type MUST be different from current type

**REQUIRED OUTPUT FORMAT (JSON):**
```json
\{
  "qa\_decisions": [
    \{
      "qa\_index": 0,
      "action": "KEEP|DELETE|TYPEFIX",
      "reason": "specific reason for action",
      "new\_type": "explicit|implicit",
      "issues": ["specific issues found"]
    \},
    \{
      "qa\_index": 1,
      "action": "KEEP|DELETE|TYPEFIX",
      "reason": "specific reason for action",
      "new\_type": "explicit|implicit",
      "issues": ["specific issues found"]
    \},
    \{
      "qa\_index": 2,
      "action": "KEEP|DELETE|TYPEFIX",
      "reason": "specific reason for action",
      "new\_type": "explicit|implicit",
      "issues": ["specific issues found"]
    \}
  ],
  "summary": \{
    "total\_qa\_pairs": 3,
    "keep\_count": 0,
    "delete\_count": 0,
    "typefix\_count": 0,
    "quality\_assessment": "high|medium|low"
  \}
\}
\end{tcblisting}

\subsubsection{Counterfactual Distractor Generation Prompt}

D-SCoRE generates counterfactual (distractor) answers using the following prompt templates, conditioned on the QA type. Placeholders such as \texttt{\{content\}} are replaced at runtime.

% Explicit version
\begin{tcblisting}{
    breakable,
    enhanced,
    boxrule=0.4pt,
    colback=gray!6,
    colframe=gray!40,
    arc=3pt,
    fontupper=\ttfamily\small,
    listing only,
    title={Explicit QA Type},
    listing options={numbers=none}
}
You are generating multiple choice distractors. Given the content, question, and correct answer, create 3 incorrect but plausible alternatives.

**Content:**
\{content\}

**Question:** \{question\}
**Correct Answer (a0):** \{correct\_answer\}

**Task**: Generate 3 plausible but INCORRECT answers for this explicit question.

**Requirements for Explicit QA Distractors:**
- Must be factually incorrect relative to the content
- Should be plausible enough to challenge readers
- Must be the same type/format as the correct answer
- Should relate to the content but be wrong details
- Avoid obvious errors (like wrong time periods for historical facts)

RESPONSE FORMAT REQUIREMENTS:
- Output MUST be valid JSON only
- No explanations, thoughts, or additional text
- Exactly 3 alternatives labeled a1, a2, a3
- Each alternative must be a complete standalone answer
- Keep similar length/style to the correct answer

Output ONLY a valid JSON object in this exact format:
\{
    "a1": "first incorrect alternative",
    "a2": "second incorrect alternative", 
    "a3": "third incorrect alternative"
\}

Generate the 3 distractors:
\end{tcblisting}

% Implicit version
\begin{tcblisting}{
    breakable,
    enhanced,
    boxrule=0.4pt,
    colback=gray!6,
    colframe=gray!40,
    arc=3pt,
    fontupper=\ttfamily\small,
    listing only,
    title={Implicit QA Type},
    listing options={numbers=none}
}
You are generating multiple choice distractors. Given the content, question, and correct answer, create 3 incorrect but plausible alternatives.

**Content:**
\{content\}

**Question:** \{question\}
**Correct Answer (a0):** \{correct\_answer\}

**Task**: Generate 3 plausible but INCORRECT answers for this implicit question.

**Requirements for Implicit QA Distractors:**
- Must be factual inaccuracies, inconsistencies with the text, or logical flaws
- Avoid answers that contradict basic facts from the content

RESPONSE FORMAT REQUIREMENTS:
- Output MUST be valid JSON only
- No explanations, thoughts, or additional text
- Exactly 3 alternatives labeled a1, a2, a3
- Each alternative must be a complete standalone answer
- Keep similar length/style to the correct answer

Output ONLY a valid JSON object in this exact format:
\{
    "a1": "first incorrect alternative",
    "a2": "second incorrect alternative", 
    "a3": "third incorrect alternative"
\}

Generate the 3 distractors:
\end{tcblisting}

\end{document}